\documentclass[runningheads]{llncs}
\usepackage{graphicx}
\usepackage[pagebackref]{hyperref}
\hypersetup{
    colorlinks=true,
    citecolor=red,
    linkcolor=americanrose,
    citecolor=myblue,
    urlcolor=myblue
}

\usepackage{tikz}
\usepackage{comment}
\usepackage{amsmath,amssymb} 
\DeclareMathOperator*{\argmax}{argmax}

\usepackage{algorithm}
\usepackage{algpseudocode}

\usepackage{fontawesome}

\usepackage{xcolor}
\definecolor{ao(english)}{rgb}{0.0, 0.5, 0.0}
\definecolor{azure(colorwheel)}{rgb}{0.0, 0.5, 1.0}
\definecolor{americanrose}{rgb}{1.0, 0.01, 0.24}
\definecolor{myred}{rgb}{0.753, 0.314, 0.275}
\definecolor{myblue}{rgb}{0.0, 0.24, 0.95}
\definecolor{applegreen}{rgb}{0.55, 0.91, 0.0}
\definecolor{tbl_gray}{gray}{0.85}

\usepackage{overpic}
\usepackage{caption}
\usepackage{subcaption}
\usepackage{wrapfig}

\usepackage[capitalize]{cleveref}
\crefname{section}{Sec.}{Secs.}
\Crefname{section}{Section}{Sections}
\Crefname{table}{Table}{Tables}
\crefname{table}{Tab.}{Tabs.}

\usepackage{setspace}

\newcommand{\mypar}[1]{\vspace{0.3em}\noindent\textbf{#1}.}
\newcommand{\revise}[1]{{\color{black}{#1}}}

\newcommand{\eg}{\emph{e.g.}}
\newcommand{\etal}{\emph{et al.}}

\newcommand{\mysection}[1]{\vspace{-0.3em}\section{#1}\vspace{-0.3em}}
\newcommand{\mysubsection}[1]{\vspace{-0.3em}\subsection{#1}\vspace{-0.3em}}

\usepackage{booktabs} 
\usepackage{array}
\usepackage{diagbox}
\usepackage{multirow}
\usepackage{colortbl} 
\usepackage{makecell}
\usepackage{arydshln} 

\newlength\savedwidth
\newcommand{\whline}[1]{\noalign{\global\savedwidth\arrayrulewidth \global\arrayrulewidth #1}%
                   \hline \noalign{\global\arrayrulewidth\savedwidth}}

\usepackage[accsupp]{axessibility}  

\newcommand{\arXiv}{}

\ifdefined \arXiv
\else
   \usepackage{ruler}
   \usepackage[width=122mm,left=12mm,paperwidth=146mm,height=193mm,top=12mm,paperheight=217mm]{geometry}
\fi

\usepackage{soul}
\definecolor{airforceblue}{rgb}{0.36, 0.54, 0.66}

\begin{document}
\pagestyle{headings}
\mainmatter
\def\ECCVSubNumber{5369}  

\title{
   Geometric Synthesis: A Free lunch for Large-scale Palmprint Recognition  Model Pretraining
}

\ifdefined \arXiv
\titlerunning{Geometric Synthesis}
\author{
   Kai Zhao\inst{1,2}\thanks{KZ is the corresponding author: \href{mailto:kz@kaizhao.net}{kz@kaizhao.net}.} \and
   Lei Shen\inst{1} \and
   Yingyi Zhang\inst{1} \and
   Chuhan Zhou \inst{1} \and
   Tao Wang \inst{1}, \\
   Ruixin Zhang \inst{1} \and
   Shouhong Ding  \inst{1} \and
   Wei Jia \inst{3} \and
   Wei Shen \inst{4}
}
\authorrunning{K. Zhao et al.}
\institute{
   Tencent Youtu Lab \and
   University of California, Los Angeles \and
   Hefei University of Technology \and
   Shanghai Jiaotong University
}
\else
\titlerunning{ECCV-22 submission ID \ECCVSubNumber} 
\authorrunning{ECCV-22 submission ID \ECCVSubNumber} 
\author{Anonymous ECCV submission}
\institute{Paper ID \ECCVSubNumber}
\fi
\maketitle

\begin{abstract}
Palmprints are private and stable information for biometric recognition.
In the deep learning era, the development of palmprint recognition is limited by the lack
of sufficient training data.
In this paper, by observing that palmar creases are the key information to deep-learning-based palmprint recognition,
we propose to synthesize training data by manipulating palmar creases.
Concretely, we introduce an intuitive geometric model which represents palmar creases
with parameterized B\'ezier curves.
By randomly sampling B\'ezier parameters, we can synthesize massive training samples of diverse identities,
which enables us to pretrain large-scale palmprint recognition models.
Experimental results demonstrate that such synthetically pretrained models
have a very strong generalization ability: they can be efficiently
transferred to real datasets, leading to significant performance improvements on palmprint recognition.
For example, under the open-set protocol, our method improves the strong ArcFace baseline
by more than 10\% in terms of TAR@1e-6.
And under the closed-set protocol, our method reduces the equal error rate (EER) by an order of magnitude.
%
%
\ifdefined \arXiv
Codes are available at \url{http://kaizhao.net/palmprint}.
\else
The code will be made openly available upon acceptance.
\fi
\end{abstract}

\mysection{Introduction}\label{sec:intro}

Palm information is privacy-by-design because the palm
pattern is concealed inside your hand,
and it is almost impossible
to be tracked by public cameras without your consent.
For its security and privacy, palmprint recognition is being adopted 
by AmazonOne for identification and payment ~\cite{amazonone}.
In contrast, the widely used face recognition system can easily track people
through public cameras without any consent.
As a result, face recognition has received widespread criticism in the last few
years due to the privacy concerns it creates~\cite{van2020ethical,gibney2020battle}.

It is precisely because of its privacy that palmprint recognition lacks a large-scale public dataset.
To the best of our knowledge, the largest open dataset for palmprint recognition
contains thousands of identities and tens of thousands of images~\cite{kumar2018toward}.
In contrast, there are a number of million-scale face recognition datasets
either based on webly collected faces~\cite{kemelmacher2016megaface,cao2018vggface2,guo2016ms}
or surveillance cameras~\cite{maze2018iarpa,whitelam2017iarpa}.
The lack of sufficient data has become the main bottleneck for palmprint recognition.
In this paper, we propose to synthesize images to augment the training set for
palmprint recognition.
\begin{figure}
   \centering
   \begin{overpic}[width=0.8\linewidth]{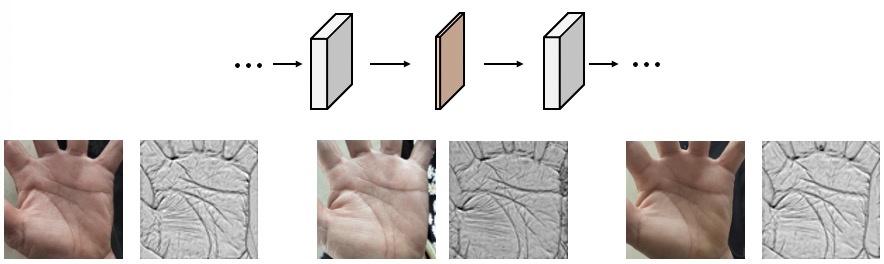}
   \end{overpic}\vspace{-8pt}
   \caption{
      Top: during training, we squeeze the intermediate features to 1-dimension for visualization.
      Middle and bottom: the input palm images and intermediate features.
      The intermediate features suggest that palmar creases are the key information to palmprint recognition.
   }\label{fig:palm-cams}\vspace{-8pt}
\end{figure}

By visualizing the intermediate features from a CNN-based palmprint recognition model,
we observe that the palmar creases play a critical role.
%
%
%
Specifically, we insert a `squeeze-and-excite' operation
into an existing CNN architecture.
The `squeeze-and-excite' first squeezes an intermediate feature map
into 1-dimension and excites it back.
Then we train this model on palmprint datasets.
Finally, we visualize the 1-dimensional feature map on test images.
As shown in~\cref{fig:palm-cams}, most of the texture and color information
are ignored by the model, and the palmar creases are largely reserved.
%
This reminds us that we may synthesize palmar creases to enrich the training
data for palmprint recognition.

With the above observation, 
we propose to synthesize training data for palmprint recognition
by manipulating palmar creases.
%
%
An intuitive and simple geometric model is proposed to synthesize palm images
by representing palmar creases with several parameterized B\'ezier curves.
The identity of each synthesized data is controlled by the parameters of B\'ezier curves,
\eg number of curves, positions of endpoints, and control points.
Our method is able to synthesize massive samples of diverse identities,
which enables us to perform large-scale pretraining on such synthetic data.
The synthetically pretrained models present promising generalization ability and can be
efficiently transferred to real datasets.

\revise{
Our method holds essential differences from other data generation methods
such as generative adversarial networks (GANs) and data augmentation.
First of all, both GANs and data augmentation rely on existing
data: either train GANs with or add modifications to existing data.
While our method creates new samples without any existing data.
Second, neither GANs nor data augmentation can create samples
of novel categories, while our method can control the category (identity)
of synthesized samples.
In addition, GANs require large amount of training data and thus
cannot substantially improve the recognition performance.
For example, ~\cite{qiu2021synface} uses GAN-synthesized samples to train face
recognition models.
However, the synthetically trained models perform worse than models that are directly
trained on dataset that is used to train GANs.
}
%

Extensive experiments under both open-set and closed-set evaluation
protocols demonstrate that our method significantly improve the performance of
strong baselines.
Additionally, we evaluate on a private million-scale data to further
confirm the scalability of our method.
%
The contributions of this paper are summarized as below:
\vspace{-6pt}
\begin{itemize}\setlength\itemsep{-0.2em}
   \item We visualize the intermediate features of CNN-based palmprint recognition
         models and observe that the palmar creases play an important role.

   \item We propose a simple yet effective model to synthesize training data by manipulating creases
         with parameterized curves.
         We pretrain deep palmprint recognition models with the synthetic data and then finetune
         them on real palmprint datasets.
   
   \item Extensive evaluation on 13 public datasets 
         demonstrates that the synthetically pretrained
         models significantly outperform their `train-from-scratch' counterparts
         and achieve state-of-the-art recognition accuracy.
   \item We test our method on a million-scale dataset,
         which is, to the best of our knowledge, the largest evaluation
         in palmprint recognition.
         The results verify the scalability of our method,
         showing its strong potential in the industry-level
         palmprint recognition.
\end{itemize}

\mysection{Related Work}\label{sec:related-work}
\mysubsection{Palmprint Recognition}
\mypar{Traditional palmprint recognition}
Traditional palmprint recognition methods in the literature can be roughly classified
into two categories:
holistic-based and local-based.
In holistic-based methods, features are extracted from the whole image
and then projected to a space of lower-dimensional to make it more compact
and discriminative.
PCA~\cite{lu2003palmprint} and its 2D variant~\cite{sang2009research}
are commonly used in this category.
Besides, independent component analysis (ICA) is also used~\cite{connie2005automated}.
PCA seeks to find uncorrelated features while ICA attempts to
find statistically independent features.
Supervised projection methods including Linear Discriminant Analysis (LDA)
and 2D-LDA~\cite{wang2006palmprint} have also been explored.
Another interesting method Locality Preserving Projection (LPP)~\cite{he2003locality}
attempts to preserve the local structure of images.
Hu~\etal~\cite{hu2007two} extend LPP to 2D
and Feng~\etal~\cite{feng2006alternative} introduce non-linear kernel to LPP.
The holistic-based methods often suffer from degradation caused by distortion,
illumination, and noise.
To overcome these issues, the palm images are firstly transformed to
another domain.
Frequency~\cite{hennings2007palmprint,li2012palmprint}, Cosine~\cite{leng2017dual,laadjel2015combining} and Radon~\cite{tamrakar2016noise}
transforms are commonly used to overcome these degradations.

Local-based methods extract local features on the image and then fuse these
features globally for recognition.
%
%
Competitive coding (CompCode)~\cite{kong2004competitive} uses 2-D Gabor filters to extract
orientation information from palm lines.
%
%
FastCompCode~\cite{zheng2015suspecting} proposes a binary code for effective representation
and matching.
Other coding-based methods include SMCC~\cite{zuo2010multiscale},
RLOC~\cite{jia2008palmprint},
ContourCode~\cite{khan2011contour},
double orientation code~\cite{fei2016double}, \etal.
%
%
%
Wu~\etal~\cite{wu2014sift} extract local SIFT features and match palm images with RANSAC.
Qian~\etal~\cite{qian2013discriminative} extract histogram of orientations.
%

\mypar{Deep learning based palmprint recognition}
Inspired by the success of deep learning in other
recognition tasks such as person re-identification and
face recognition~\cite{deng2019arcface,zhao2019regularface}, many researchers attempt to use deep learning
technologies for palmprint recognition.
Dian~\etal~\cite{dian2016contactless} use the AlexNet as the feature extractor
and match palm images with Hausdorff distance.
Svoboda~\etal~\cite{svoboda2016palmprint} train CNNs with a novel loss function related to
the d-prime index.
Recently, margin-based loss functions have been proven to be effective for face recognition.
The large margin loss~\cite{zhong2019centralized}
and additive angular margin loss~\cite{zhang2020towards} are
introduced to palmprint recognition and impressive performance has been achieved.
Graph neural networks are also used for palmprint recognition to model the geometric structure of palmprints~\cite{shao2019few}.
Shao~\etal~\cite{shao2021deep} combine deep learning with hash coding
to build efficient palmprint recognition models.
Different from these studies that introduce new architectures
or loss functions,
our proposed method focuses on synthesizing training data for deep palmprint recognition.

\mysubsection{Data Synthesis for Deep Models}
Data synthesis aims at synthesizing training data to reduce
the cost of data acquisition and labeling.
Gaidon~\etal~\cite{gaidon2016virtual} render the street views
to pretrain deep models for object tracking.
Tremblay~\etal~\cite{tremblay2018training} render similar sceens for
object detection.
Yao~\etal~\cite{yao2020simulating} use a graphic engine to simulate a large amount of training data
for autonomous driving.
%
%
Varol~\etal~\cite{varol2017learning} synthesize images from
3D sequences of human motion capture data for human pose estimation.
Sharingan~\cite{pnvr2020sharingan} Combines synthetic and real data for unsupervised geometry estimation.
Baek~\etal~\cite{baek2018augmented} synthesize depth maps with generative adversarial networks~\cite{NIPS2014_5ca3e9b1}
for depth-based human pose estimation.
To reduce the gap between synthetic and natural images,
Shrivastava~\etal~\cite{shrivastava2017learning} proposed the Simulated+Unsupervised learning paradigm
and
Chen~\etal~\cite{chen2020automated} propose a layer-wise learning rate
selection method to improve the synthetic-to-real generalization performance.
All these methods synthesize samples of \emph{existing} and \emph{known}
categories,
while our proposed method aims at generating samples for  \emph{novel} categories
and augmenting the training identities for palmprint recognition.

\begin{figure}
    \centering
    \begin{subfigure}[b]{0.45\textwidth}
        \centering
        \includegraphics[width=0.6\textwidth]{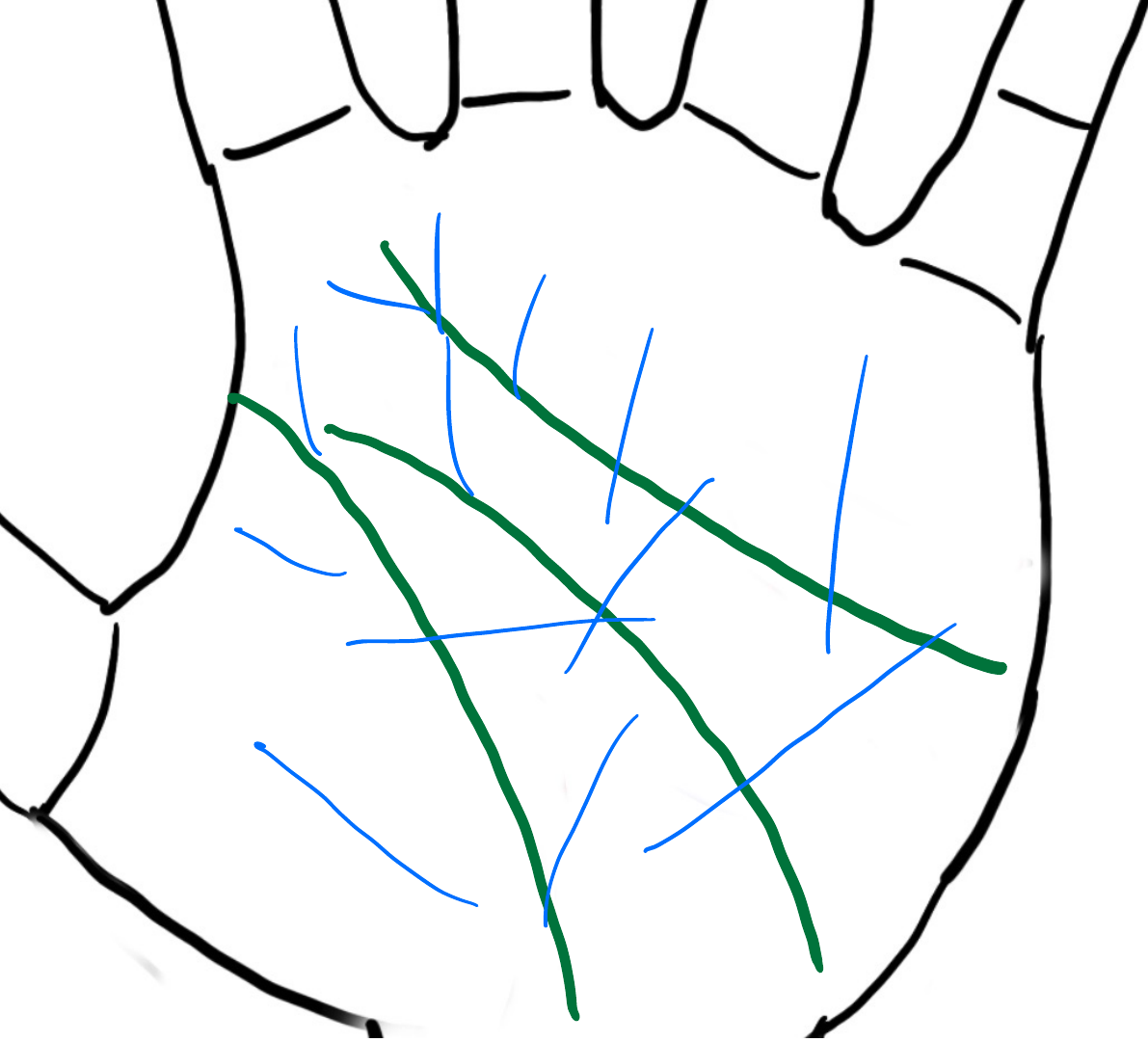}
        \caption{
         A left palm with 3 principal lines ({\color{ao(english)}{green}})
         and several wrinkles ({\color{azure(colorwheel)}{blue}}).
        }
        \label{fig:palm-sketch}
    \end{subfigure}
    \hfill
    \begin{subfigure}[b]{0.45\textwidth}
        \centering
        \includegraphics[width=0.6\textwidth]{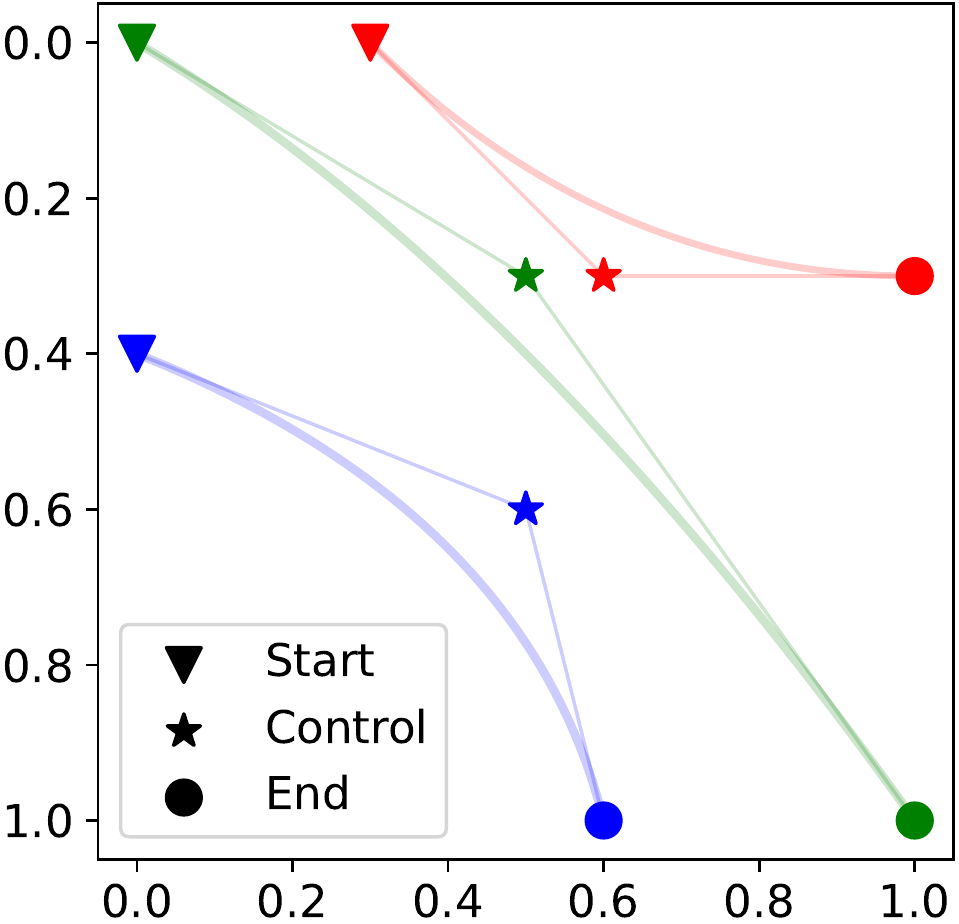}
        \caption{
         Control points (
         Start (\textcolor{black}{$\blacktriangledown$}),
         control (\textcolor{black}{$\star$})
         and end ({\tiny{\faCircle}}))
         of principal lines.
        }
        \label{fig:bezier-curves}
    \end{subfigure}
    \vspace{-2pt}
    \caption{
        An example hand (left) and control points of principal lines (right).
    }\vspace{-10pt}
\end{figure}

\mysection{Methodelogy}\label{sec:method}

As illustrated in~\cref{fig:palm-sketch}, the palmprints are roughly composed of several (usually 3$\sim$5) principal lines
and a number of thin wrinkles.
To imitate the geometric appearance of palmprints, we use the B\`ezier curves
to parameterize the palmar creases.
Specifically, we use several B\`ezier curves to represent the principal lines and
the wrinkles.
For simplicity, we use second-order B\`ezier curves with three parametric points in
a 2D plane, a control point, a start point, and an end point.
~\cref{fig:bezier-curves} gives an example of the parametric points 3 principal lines of a left hand.
Next, we will take the left hand as an example to detailedly illustrate how we determine the parameters of B\`ezier curves,
and the case for the right hand can be regarded as the mirror of the left hand.

\mysubsection{Palmar creases with B\`ezier Curves}
Let $N$ and $S$ be the number of total identities and number of samples for each identity,
we will generate $N\times S$ samples in total.
\vspace{-10pt}
\begin{figure}[!htb]
   \centering
   \begin{subfigure}[b]{0.45\textwidth}
       \centering
       \includegraphics[width=0.6\textwidth]{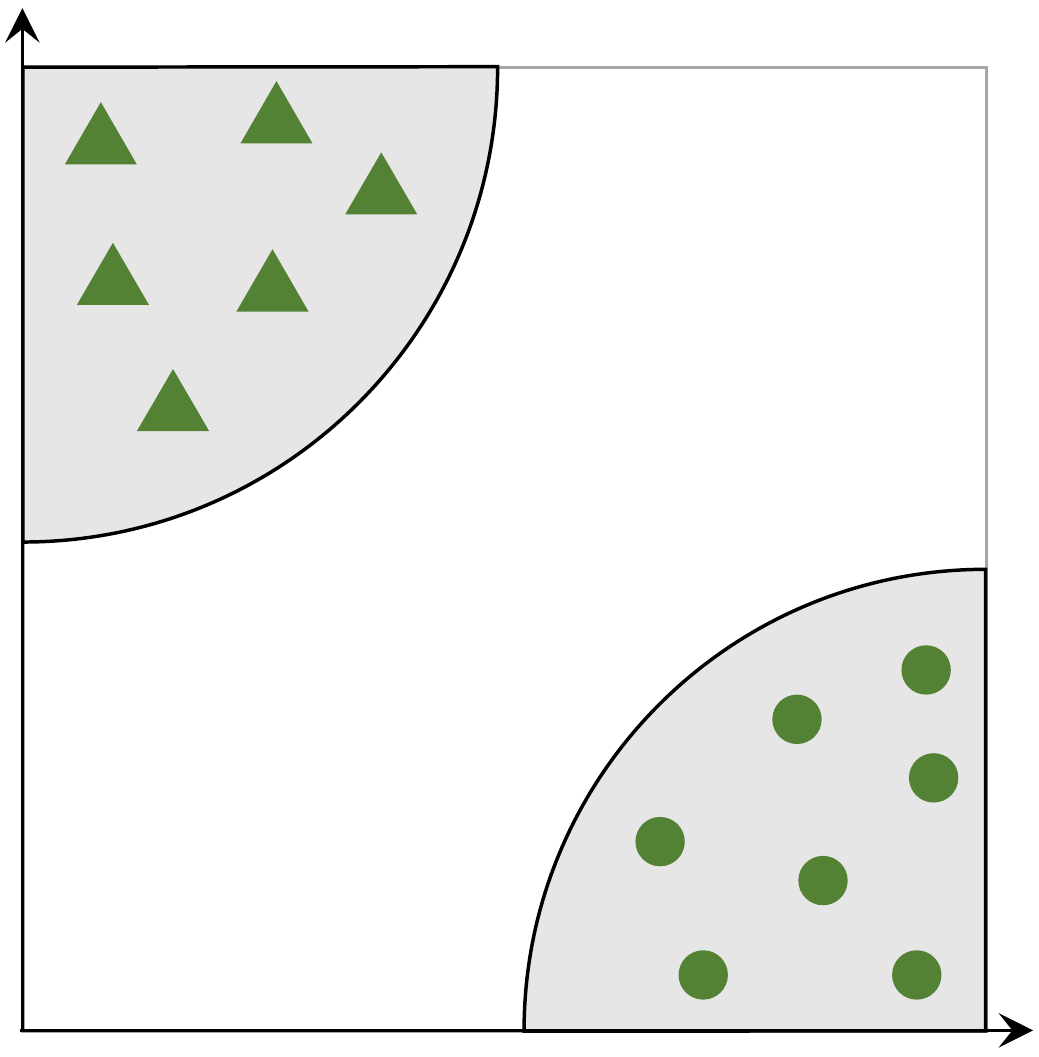}
       \caption{
        Start and end points of principal lines are sampled from top-right and bottom-left
        corners.
       }
       \label{fig:start-end-points}
   \end{subfigure}
   \hfill
   \begin{subfigure}[b]{0.45\textwidth}
       \centering
       \includegraphics[width=0.6\textwidth]{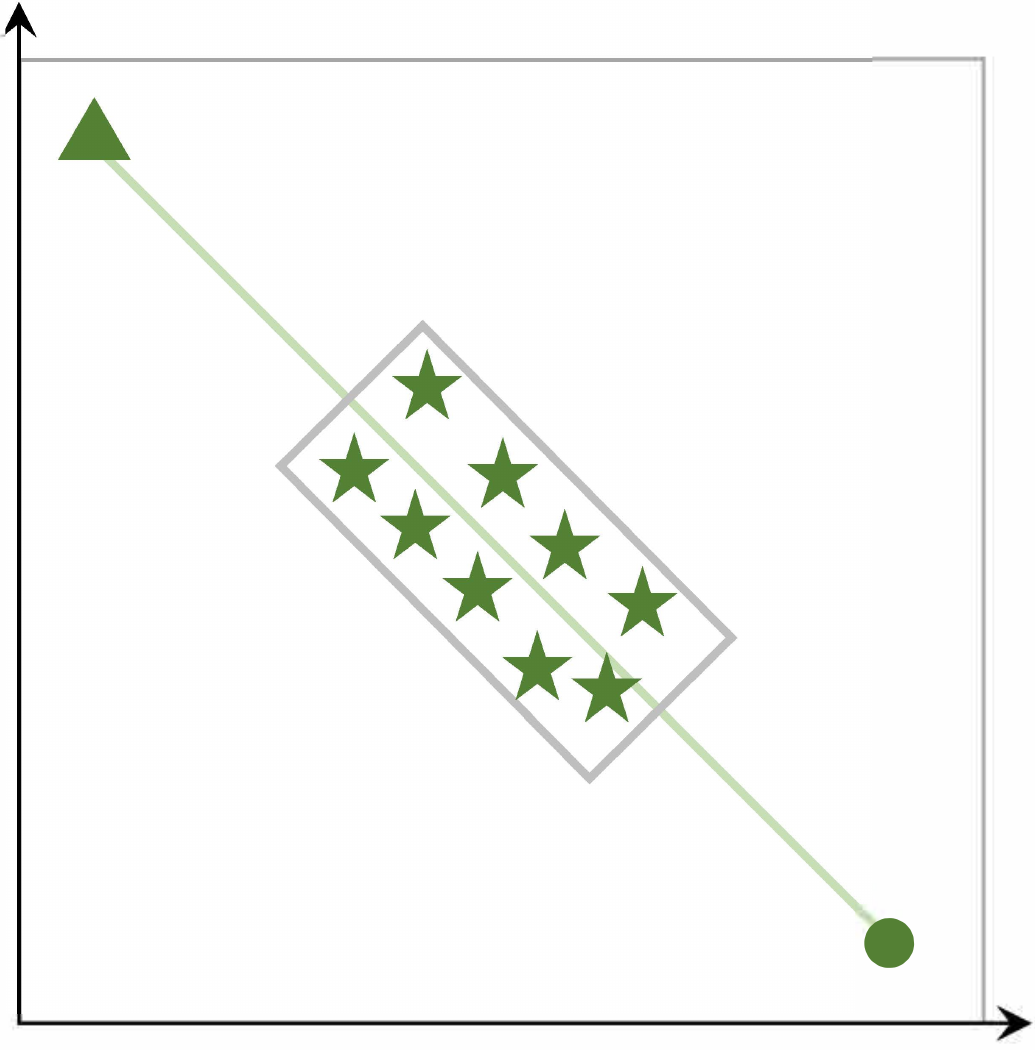}
       \caption{
        The control point is sampled from a rectangle
        that is parallel to the line connecting starting and end points.
       }
       \label{fig:control-points}
   \end{subfigure}
   \vspace{-10pt}
   \caption{
       Start ($\blacktriangledown$), end ({\tiny{\faCircle}}) and control ({\large{$\star$}}) points.
   }
\end{figure}\vspace{-10pt}
For each identity, we synthesize $m$ principal lines and $n$ wrinkles,
where $m$ and $n$ are sampled from uniform distributions: $m \sim U(2, 5)$ and $n \sim U(5, 20)$.
\revise{
Take left hand as an example, the starting and end points of principal lines are randomly
sampled from top-left and bottom-right corner of the plane, as shown in~\cref{fig:start-end-points}.
While the start and the end points of wrinkles are randomly sampled from the whole plane.
Then, given the starting and end points, the control point is sampled from a rectangle that is parallel
to the line connecting two points, as shown in~\cref{fig:control-points}.
}

%
%
%


\mysubsection{Parameters of B\`eziers}
We first determine the number of principal lines and wrinkles
for each identity, and then randomly sample
start, end, and control points for each crease.


\mypar{principal lines}
For each identity, we sample $m$ principal lines starting from the
top-left corner to the bottom-right corner.
For left palms, the start/end points for each principal line are sampled from the top-left and bottom-right
corner.
Given the start/end points of a B\`ezier, its control point is randomly
sampled from a rectangle area in the middle of the line connecting start and end points.
The details about the synthesis of principle lines are illustrated in~\cref{appendix:synthesize-details}.

\mypar{wrinkles}
We generate $n = 5\sim 15$ wrinkles for each identity.
We do not restrict the directions of wrinkles
and their start, end, and control points are randomly sampled from the whole plane:
$$
Q = \text{random}(0, 1, \text{size}=(n, 3, 2)).
$$

\mysubsection{Within-identity Diversity}
We enhance the within-identity diversity of synthesized samples in two aspect:
1) we add small random noices to the parameters so that each sample is a little different from others;
2) we use a randomly selected natural image as the background of the synthesized sample.

\vspace{-15pt}
\subsubsection{Random noice.}
Given parameters of a specific identity, we add small noises
to $P$ and $Q$ to synthesize diverse samples.
Formally, the parameters for the $j$-th sample of identity $i$
are:
\begin{equation}
   \begin{split}
      P^i_j &= P^i + N_p \\
      Q^i_j &= Q^i + N_q,
   \end{split}
\end{equation}
where $N_p \sim \mathcal{N}(\mu,\, 0.04)$ and $ N_q \sim \mathcal{N}(\mu,\, 0.01)$
are small gaussian noises.
Each crease is rendered with a random color $c$ and stroke width $w$.

\vspace{-15pt}
\subsubsection{Random Background.}
For each sample we select a random image from the imagenet~\cite{deng2009imagenet} dataset
as the background of the synthesized sample.

Finally, sample $S^i_j$ is synthesized with:
$$
S^i_j = \text{synthesize}(P^i_j, Q^i_j, c, w, I)
$$
A overall algorithmic pipeline is illustrated in~\cref{appendix:algorithm}
and some synthetical samples can be found in~\cref{appendix:example}.

\mysection{Experimental Settings}\label{sec:exp-settings}
In this section, we introduce the detailed experimental
settings including data preparation and evaluation protocols.
\revise{
Our experiments are mainly based on the ArcFace~\cite{deng2019arcface}, a strong baseline for
palm recognition~\cite{zhang2020towards}.
During training, we use the ArcFace loss as supervision.
During testing, we extract 512 dimensional features for each sample and the cosine similarity
is used as the distance measurement.
}

\begin{figure}[!htb]
   \centering
   \begin{overpic}[width=1\linewidth]{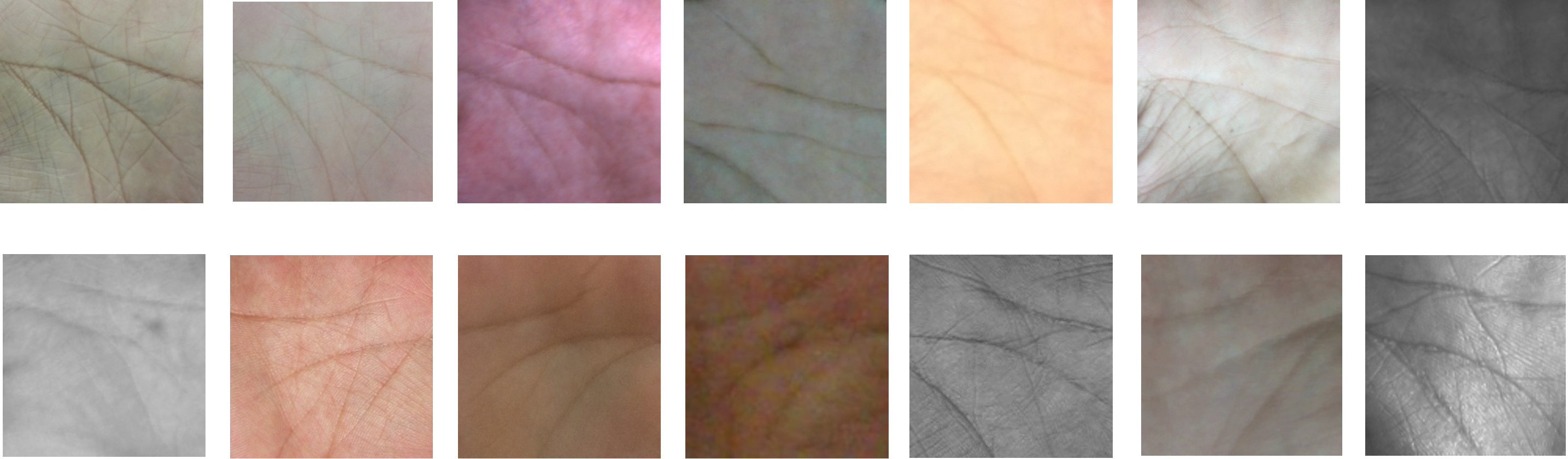}
      \put(3.8,14.2){\footnotesize{MPD}}
      \put(18.8,14.2){\footnotesize{MPD}}
      \put(33,14.2){\footnotesize{TCD}}
      \put(45,14.2){\footnotesize{XJTU\_A}}
      \put(58,14.2){\footnotesize{XJTU\_UP}}
      \put(76,14.2){\footnotesize{IITD}}
      \put(87,14.2){\footnotesize{CASIA-MS}}
      \put(3.8,-2){\footnotesize{CASIA}}
      \put(18.8,-2){\footnotesize{COEP}}
      \put(32,-2){\footnotesize{MOHI}}
      \put(46,-2){\footnotesize{WEHI}}
      \put(58.5,-2){\footnotesize{PolyU-MS}}
      \put(75.5,-2){\footnotesize{GPDS}}
      \put(86,-2){\footnotesize{PolyU 2d+3d}}
   \end{overpic}\vspace{2pt}
   \caption{
      Example ROIs of different datasets.
   }\label{fig:example-rois}
\end{figure}
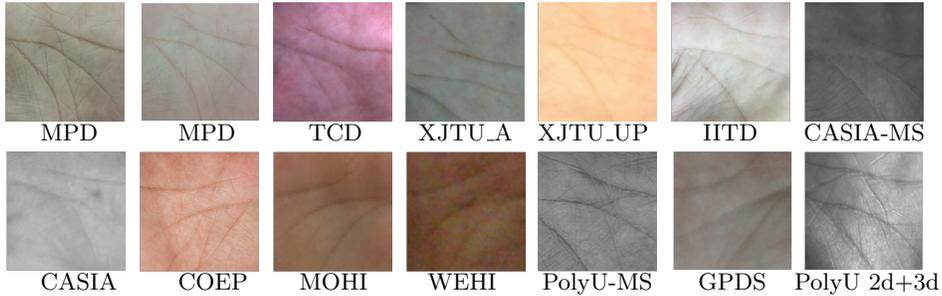

\mysubsection{Datasets and Data Preprocessing}
\mypar{Datasets}
We use 13 public datasets in our experiments.
The statistical information of these datasets is summarized in~\cref{tab:dataset-stats}
and example ROIs of these datasets are shown in~\cref{fig:example-rois}.
\begin{table}[!h]
   \centering
   \vspace{-4pt}
   \resizebox{1\linewidth}{!}{
   \renewcommand{\arraystretch}{1.1}
   \begin{tabular}{lccc|lcccc}
      Name & \#IDs & \#Images & Device & Name & \#IDs & \#Images & Device \\
      \whline{1.2pt}
      MPD~\cite{zhang2020towards}             & 400 & 16,000  & Phone & COEP~\cite{ceop}                        & 167 & 1,344 & Digital camera \\
      XJTU\_UP~\cite{shao2020effective}       & 200 & 30,000  & Phone & TCD~\cite{zhang2020towards}             & 600 & 12,000  & Contactless\\
      MOHI~\cite{hassanat2015new}             & 200 & 3,000 & Phone & IITD~\cite{kumar2008incorporating}      & 460 & 2,601 & Contactless \\
      GPDS~\cite{ferrer2011bispectral}        & 100 & 2,000 & Web cam & CASIA~\cite{sun2005ordinal}                 & 620 & 5,502 & Contactless\\
      WEHI~\cite{hassanat2015new}             & 200 & 3,000 & Web cam & PolyU-MS~\cite{zhang2009online}         & 500 & 24,000 & Contactless \\
      PolyU(2d+3d)~\cite{kanhangad2010contactless} 
                                              & 400 & 8,000 & Web cam & CASIA-MS~\cite{hao2008multispectral}    & 200 & 7,200 & Contactless \\
      XJTU\_A~\cite{shao2020effective}        & 114 & 1,130  & CMOS camera &  \\
      \whline{0.1pt}
   \end{tabular}}
   \caption{Statistics of the 13 public palmprint datasets.}
   \label{tab:dataset-stats}
\end{table}

The images in CASIA-MS~\cite{zhang2009online} dataset are captured with multi-spectral devices and
we only use visible spectra images.
We remove the overlapped identities in MPD~\cite{zhang2020towards} and TCD~\cite{zhang2020towards} datasets.
Finally, there are 3,268 identities and 59,162 images used in our experiments.

\mypar{ROI extraction}
We follow the protocol of~\cite{zhang2019pay} for ROI extraction.
Given a
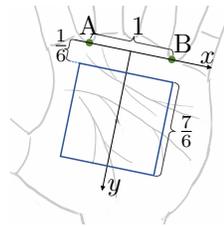
\begin{wrapfigure}{r}{0.4\textwidth}
   \centering
   \vspace{-10pt}
   \begin{overpic}[width=0.6\linewidth]{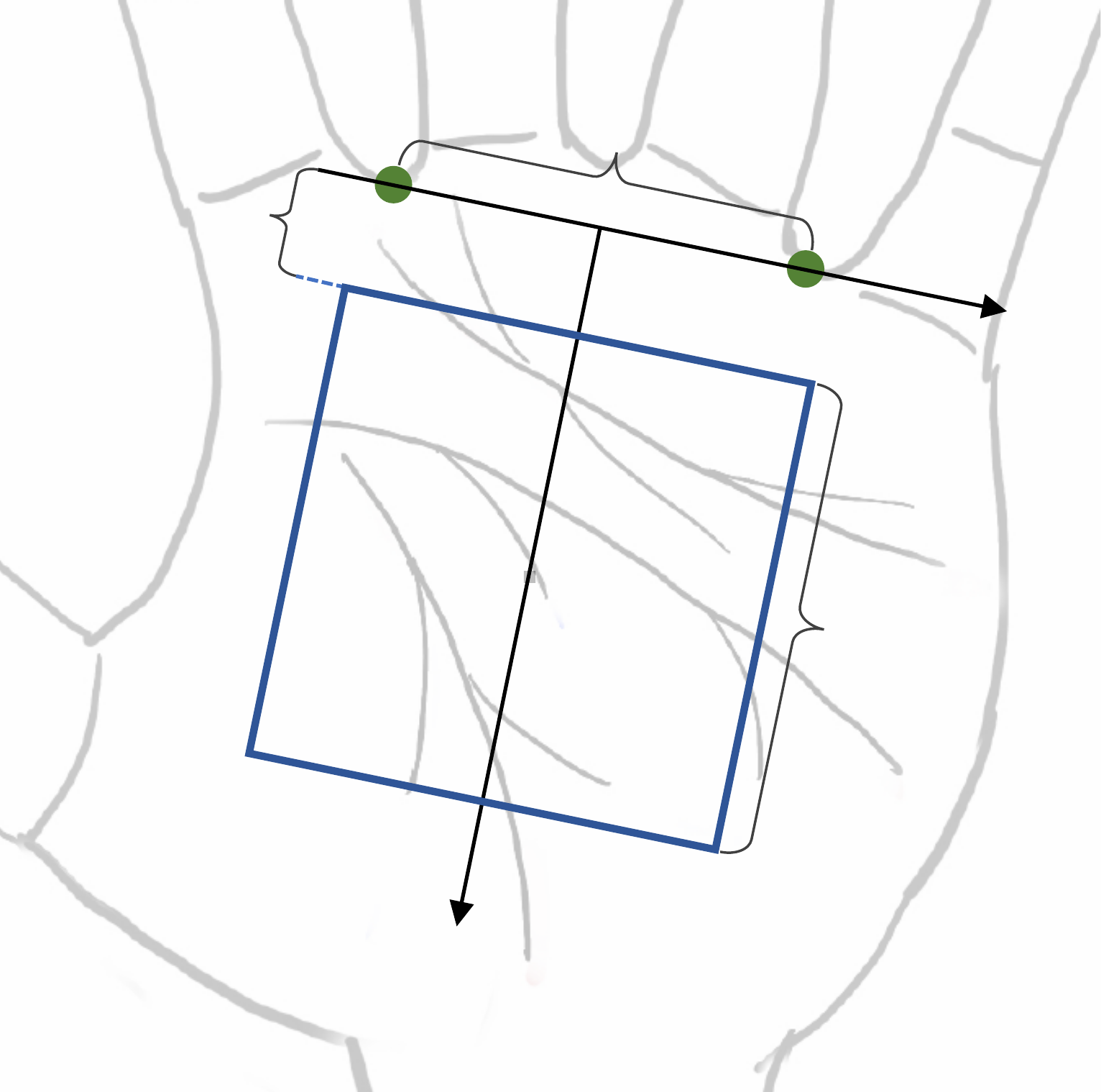}
      \put(76,39){$\frac{7}{6}$}
      \put(55,86){$1$}
      \put(19,78){$\frac{1}{6}$}
      \put(31,86){A}
      \put(74,78){B}
      \put(86,73){$x$}
      \put(44,15){$y$}
   \end{overpic}\vspace{-4pt}
   \caption{
      ROI extraction of a left hand.
   }\label{fig:roi-extraction}
   \vspace{-24pt}
\end{wrapfigure}
palm image, we first detect two landmarks and then crop  the center area of
the palm according to the landmarks.
~\cref{fig:roi-extraction} illustrates the landmarks (A and B) and ROI of the left hand.
As shown in~\cref{fig:roi-extraction},
we use the intersection of the index finger and little finger as the first landmark (A),
and the intersection of the ring finger and middle finger as the second landmark (B).

Then we set up a coordinate where $\vec{AB}$ is the $x$-axis
and its perpendicular is the $y$-axis.
Suppose $|\vec{AB}| = 1$ is the unit length, we crop a square with side length $7/6$
as the ROI.
The ROI of the right hand can be extracted similarly.
Some example ROIs used in our experiments are shown in~\cref{fig:example-rois}.

\mysubsection{Open-set Protocol}\label{sec:exp-setting-openset}
\mypar{Dataset split}
For the open-set protocol, we select part of identities from each
\begin{wraptable}{r}{0.5\textwidth}
   \centering
   \vspace{-15pt}
   \resizebox{1\linewidth}{!}{
   \begin{tabular}{l|ccccc}
      Split & mode & \#IDs & \#Images \\
      \whline{1.2pt}
      \multirow{2}{*}{train:test = 1:1} & train & 1,634 & 29,347 \\
      & test & 1,632 & 29,815 \\
      \multirow{2}{*}{train:test = 1:3} & train & 818 & 14,765 \\
      & test & 2,448 & 44,397 \\
   \end{tabular}
   }
   \vspace{-7pt}
   \caption{Training/test splits of the open-set protocol.}
   \label{tab:split-stats}
   \vspace{-15pt}
\end{wraptable}
dataset
and combine them as a large training set,
and the other identities are merged as a large test set.
We test two different split settings.
In the first setting, half of the identities are used for training, and half
are used for testing.
In the second setting, 1/4 of the identities are used for training and
others for testing.
The number of samples and identities in the two splits are summarized in~\cref{tab:split-stats}.

\mypar{Evaluation}
The performance under the open-set protocol is evaluated in terms of TAR@FAR,
where TAR and FAR stand for `true accept rate' and `false accept rate', respectively.
Specifically, given several test images,
we randomly sample several positive pairs where the two samples share the same identity,
and negative pairs whose samples are from distinct identities.
Let $p^+, p^-$ be the positive/negative pairs and $sim(p)$ be the similarity
between a pair of samples.
We first fix the FAR and then calculate a proper threshold $\tau$ from negative
pairs, finally we compute TAR using that threshold on the positive pairs.

Take FAR=1e-3 as an example, we can search for a threshold $\tau$ on the negative pairs
satisfying:
$$
\text{FAR} = 10^{-3} = \frac{|\{p^- \ \ | \ \ sim(p^-) > \tau\}|}{|\{p^-\}|}.
$$
With the threshold, we then calculate the TAR on the positive pairs:
$$
\text{TAR} = \frac{|\{p^+ \ \ | \ \ sim(p^+) > \tau\}|}{|\{p^+\}|}.
$$
Accordingly, we can calculate TAR under various FARs.
In our experiments, we report the performance under FAR=1e-3, 1e-4, 1e-5, 1e-6.

\mysubsection{Closed-set Protocol}\label{sec:exp-setting-closedset}
The closed-set experiments are conducted on five datasets:
CASIA, IITD, PolyU, TCD, and MPD.
We perform 5-fold cross-validation on each dataset
and report the average performance.
The experiments are conducted individually on the five datasets.
We use top-1 accuracy, EER to evaluate the performance
of closed-set palmprint recognition.
To compute the top-1 accuracy, we randomly select one sample from each identity as the \emph{registry}
and other samples are \emph{queries}.
Let $\mathcal{R}=\{ r_i\}$ be the set of registries and $\mathcal{Q}=\{ q_j\}$ the query set,
and $sim(q_i, r_j)$ is the similarity between two samples.
$y(\cdot)$ tells the identity label of a sample.
The successfully matched queries are these queries that are of the same identity
with their nearest registries:
\begin{equation*}
   \mathcal{Q}^+ = \Big\{q_i, | \ y\Big(\argmax_{r_j \in \mathcal{R}} \ \ sim(q_i, r_j)\Big) = y(q_j) \Big\}
\end{equation*}
Finally, the top-1 accuracy is the number of successfully matched queries divided
by the total number of queries: $acc = |\mathcal{Q}^+| / |\mathcal{Q}|$.
The EER is a point where FAR (False Acceptance Rate) and FRR (False Rejection Rate) intersect.

\mysection{Experimental Results}\label{sec:exp-results}
In this section, we first compare our method with other
traditional and deep learning based palmprint recognition methods.
Then we conduct ablation studies to show the contribution of each component in our
method.

\mysubsection{Implementation Details}\label{sec:impl-details}
We implement our method with the PyTorch~\cite{paszke2019pytorch} framework.
Two backbone networks, ResNet50~\cite{he2016deep}
and MobileFaceNet~\cite{chen2018mobilefacenets}, are used in our experiments.
The B\`ezier curves are generated with an opensource
package\footnote{\url{https://bezier.readthedocs.io/}}.

\mypar{Data Synthesizing}
By default, the synthetic dataset contains 4,000 identities, and each identity
contains 100 samples.
The size of the synthetic image is $224\times 224$.
The stroke width $w$ for principal lines and wrinkles are randomly
selected from $1.5\sim 3$ and $0.5\sim 1.5$, respectively.
We randomly blur the synthetic images using a gaussian kernel
to improve generalization.

\mypar{Model Training}
\revise{For our proposed method,}
we first pretrain models on synthesized data for 20 epochs and
then finetune on real palmprint datasets for 50 epochs.
\revise{For the baseline, we directly train the models on real datasets for 50 epochs.}
We use the cosine annealing learning rate scheduler with a warmup start.
The maximal learning rate for pretraining is 0.1 and 0.01 for finetune.
All models are trained with mini-batch SGD algorithm.
The momentum is 0.9 and weight decay is set to 1e-4.
We use the \emph{additive angular margin loss} (ArcFace~\cite{deng2019arcface})
with margin $m=0.5$ and scale factor $s=48$.
Besides, we linearly warm up the margin from 0 in the first epoch to improve stability.
We use 4 GPUs to run all training experiments and each GPU process 32 images
in a batch, in total the effective batchsize is 128.

\begin{figure}[!htb]
   \centering
   \begin{overpic}[width=0.8\linewidth]{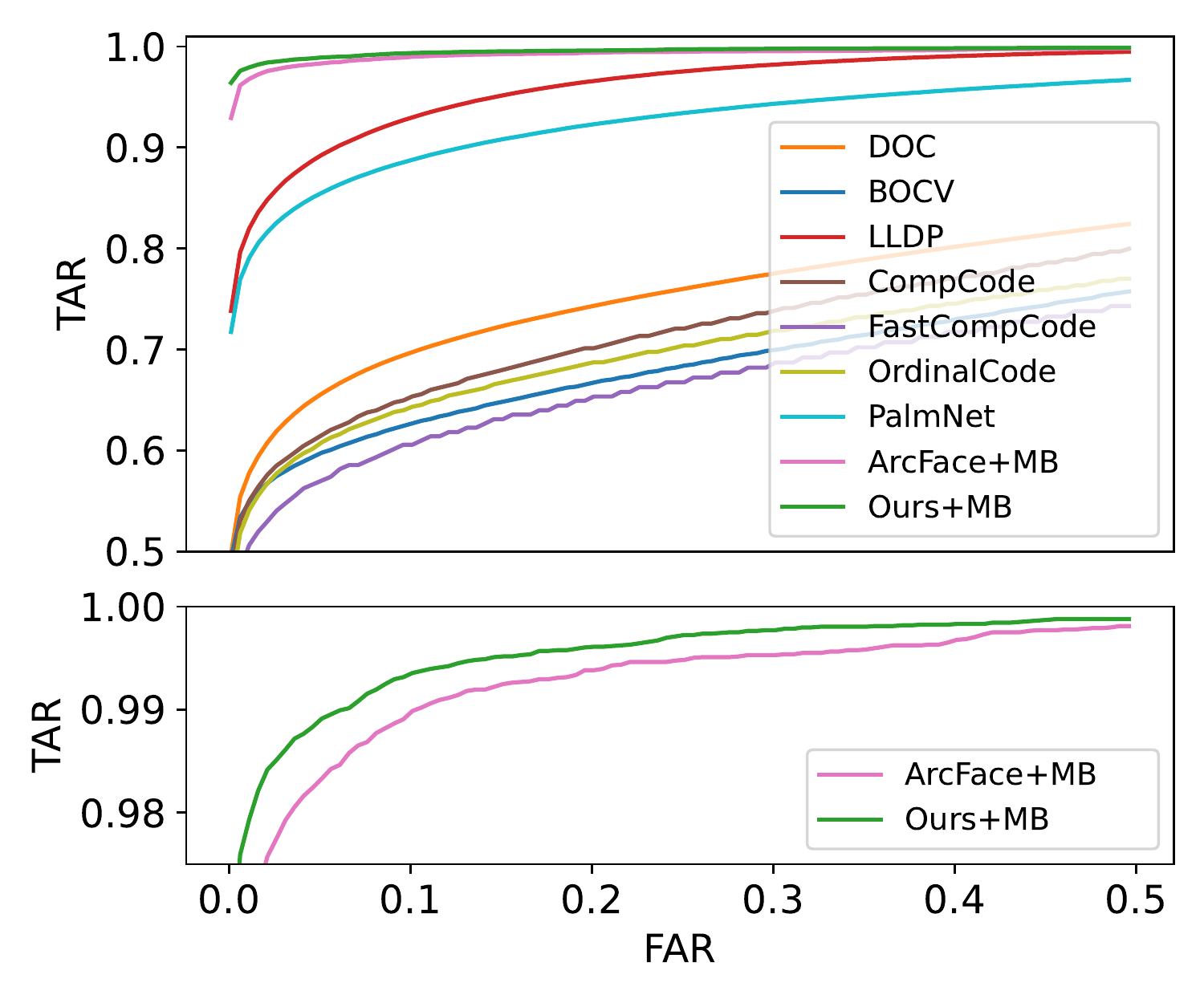}
      \put(78,69.9){\footnotesize{~\cite{fei2016double}}} 
      \put(79,66.2){\footnotesize{~\cite{guo2009palmprint}}} 
      \put(78,62.5){\footnotesize{~\cite{luo2016local}}} 
      \put(87,58.8){\footnotesize{~\cite{kong2004competitive}}} 
      \put(90.5,55.1){\footnotesize{~\cite{zheng2015suspecting}}} 
      \put(87.5,51.4){\footnotesize{~\cite{sun2005ordinal}}} 
      \put(83,47.7){\footnotesize{~\cite{genovese2019palmnet}}} 
      \put(88,44){\footnotesize{~\cite{deng2019arcface}}} 
      \put(91,18){\footnotesize{~\cite{deng2019arcface}}} 
   \end{overpic}
   \vspace{-18pt}
   \caption{
      FAR \emph{v.s.} TAR curves of various methods under the open-set 1:1 settings.
      The ArcFace and our method are based on the MobileFaceNet backbone.
   }\label{fig:roc}
\end{figure}
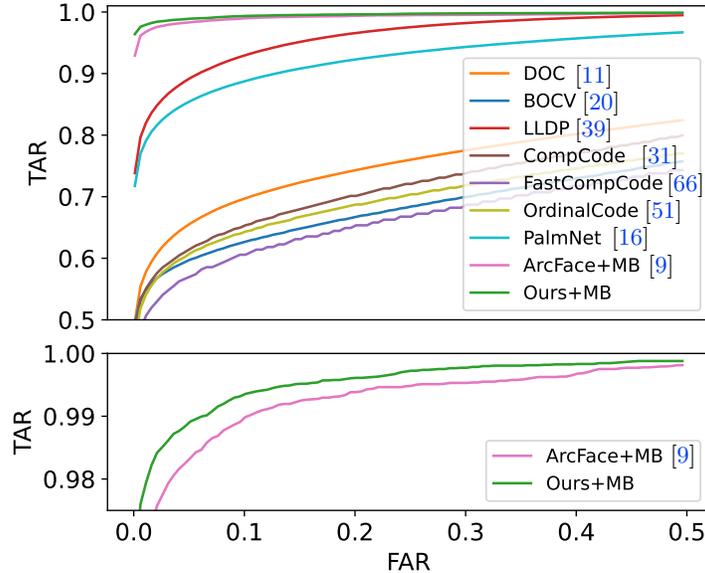

\mysubsection{Open-set Palmprint Recognition}\label{sec:open-set}
We first test our method under the "open-set" protocol.
Details about the "open-set" protocol can be found in~\cref{sec:exp-setting-openset}.
We test our method under two different training test ratios: 1:1 and 1:3,
quantitative results are in~\cref{tab:openset}.
The TAR \emph{v.s.} FAR curves of the 1:1 setting are in~\cref{fig:roc}.
\begin{table}[!htb]
   \centering
   \renewcommand{\arraystretch}{1.3}
   \resizebox{1\linewidth}{!}{
   \begin{tabular}{lc|cccc|ccccc}
   \multirow{3}{*}{Method} & \multirow{3}{*}{Backbone} & \multicolumn{4}{c|}{train  :test = 1 : 1} & \multicolumn{4}{c}{train : test = 1 : 3} \\
    &  & \makecell{TAR@ \\ 1e-3} & \makecell{TAR@ \\ 1e-4} & \makecell{TAR@ \\ 1e-5} & \makecell{TAR@ \\ 1e-6} &
                        \makecell{TAR@ \\ 1e-3} & \makecell{TAR@ \\ 1e-4} & \makecell{TAR@ \\ 1e-5} & \makecell{TAR@ \\ 1e-6} \\
   \whline{1.5pt}
   CompCode~\cite{kong2004competitive}     & N/A & 0.4800 & 0.4292 & 0.3625 & 0.2103 & 0.4501 & 0.3932 & 0.3494 & 0.2648 \\
   FastCompCode~\cite{zheng2015suspecting}  & N/A & 0.4243 & 0.3649 & 0.1678 & 0.2103 & 0.4188 & 0.3568 & 0.3100 & 0.2748 \\
   LLDP~\cite{luo2016local}                & N/A & 0.7382 & 0.6762 & 0.5222 & 0.1247 & 0.7372 & 0.6785 & 0.6171 & 0.2108 \\
   Ordinal Code~\cite{sun2005ordinal}      & N/A & 0.4628 & 0.4074 & 0.3462 & 0.1993 & 0.4527 & 0.3975 & 0.3527 & 0.2422 \\
   BOCV~\cite{guo2009palmprint}              & N/A & 0.4930 & 0.4515 & 0.3956 & 0.2103 & 0.4527 & 0.3975 & 0.3527 & 0.2422 \\
   RLOC~\cite{jia2008palmprint}            & N/A & 0.6490 & 0.5884 & 0.4475 & 0.1443 & 0.6482 & 0.5840 & 0.5224 & 0.3366 \\
   DOC~\cite{fei2016double}                & N/A & 0.4975 & 0.4409 & 0.3712 & 0.1667 & 0.4886 & 0.4329 & 0.3889 & 0.2007 \\
   PalmNet~\cite{genovese2019palmnet}      & N/A & 0.7174 & 0.6661 & 0.5992 & 0.1069 & 0.7217 & 0.6699 & 0.6155 & 0.2877 \\
   \whline{0.1pt}
   C-LMCL~\cite{zhong2019centralized}      & MB  & 0.9290 & 0.8554 & 0.7732 & 0.6239 & 0.8509 & 0.7554 & 0.7435 & 0.5932 \\
   ArcFace~\cite{deng2019arcface}          & MB  & 0.9292 & 0.8568 & 0.7812 & 0.7049 & 0.8516 & 0.7531 & 0.6608 & 0.5825 \\
   ArcFace+Ours~\cite{deng2019arcface}     & MB  & \textbf{0.9640} & \textbf{0.9438} & \textbf{0.9102} & \textbf{0.8437} &
                                                   \textbf{0.9407} & \textbf{0.8861} & \textbf{0.7934} & \textbf{0.7012} \\
   \whline{0.1pt}
   C-LMCL~\cite{zhong2019centralized}      & R50 & 0.9545 & 0.9027 & 0.8317 & 0.7534 & 0.8601 & 0.7701 & 0.6821 & 0.6254 \\
   ArcFace~\cite{deng2019arcface}          & R50 & 0.9467 & 0.8925 & 0.8252 & 0.7462 & 0.8709 & 0.7884 & 0.7156 & 0.6580 \\
   ArcFace+Ours~\cite{deng2019arcface}     & R50 & \textbf{0.9671} & \textbf{0.9521} & \textbf{0.9274} & \textbf{0.8956} &
                                                   \textbf{0.9424} & \textbf{0.8950} & \textbf{0.8217} & \textbf{0.7649} \\
   \end{tabular}
   }
   \vspace{4pt}
   \caption{
      Quantitative performance under the open-set protocol where the performance
      are evaluated in terms of TAR@FAR.
      `MB' represents the MobileFaceNets~\cite{chen2018mobilefacenets} backbone
      and `R50' is resnet50.
   }
   \label{tab:openset}
\end{table}

As shown in~\cref{tab:openset}, since traditional methods do not rely on training
data, they behave similar performance under 1:1 and 1:3 settings,
and deep learning based methods perform much better under the 1:1 setting
than under the 1:3 setting.
Among all traditional methods, LLDP~\cite{luo2016local}
performs the best.
Deep Learning based methods~\cite{genovese2019palmnet,zhong2019centralized,deng2019arcface}
significantly outperform traditional methods,
and margin-based methods, \eg C-LMCL~\cite{zhong2019centralized} and ArcFace~\cite{deng2019arcface},
present superior performance.
Our proposed method remarkably improves the ArcFace baseline and
achieves state-of-the-art performance under both 1:1 and 1:3 settings.
Under the 1:3 setting, our performance even exceeds the performance of ArcFace under the 1:1 setting.

\mysubsection{Closed-set Palmprint Recognition}\label{sec:exp-closed-set}
Here we report quantitative results of our method as well as other methods
under the closed-set protocol.
Our experiments are conducted on five datasets, and the performance
is evaluated in terms of top-1 accuracy and EER.
Detailed setting about the experiments was described in~\cref{sec:exp-setting-closedset}.

As shown in~\cref{tab:closed-set}, though the results on the closed-set protocol are nearly saturated,
our method still improves the baseline with a clear margin,
advancing the top-1 accuracies to nearly 100\%.
Besides, our method significantly decreases the EER to an unprecedented level of 1e-3,
surpassing all existing methods.
\begin{table}[!htb]
   \centering
   \resizebox{1\linewidth}{!}{
   \newcommand{\CC}[1]{\cellcolor{gray!#1}}
   \renewcommand{\arraystretch}{1.1}
   \begin{tabular}{lccccc}
      Method & CASIA & IITD & PolyU & TCD & MPD \\
      \whline{1.2pt}
      CompCode \cite{kong2004competitive}  & 79.27 / 1.08 & 77.79 / 1.39 & 99.21 / 0.68  & - / -        & - / -        \\
      Ordinal Code \cite{sun2005ordinal}           & 73.32 / 1.75 & 73.26 / 2.09 & 99.55 / 0.23  & - / -        & - / -        \\
      DoN \cite{zheng20163d}               & 99.30 / 0.53 & 99.15 / 0.68 & 100.0 / 0.22  & - / -        & - / -        \\
      \whline{0.1pt}
      PalmNet \cite{genovese2019palmnet}   & 97.17 / 3.21 & 97.31 / 3.83 & 99.95 / 0.39  & 99.89 / 0.40 & 91.88 / 6.22 \\
      FERNet \cite{matkowski2019palmprint} & 97.65 / 0.73 & 99.61 / 0.76 & 99.77 / 0.15  & 98.63 / -    & - / -        \\
      DDBC \cite{fei2019learning}          & 96.41 / -    & 96.44 / -    & -             & 98.73 / -    & - / -        \\
      RFN \cite{liu2020contactless}        & - / -        & 99.20 / 0.60 & - / -         & - / -        & - / -        \\
      C-LMCL~\cite{zhong2019centralized}   & - / -        & - / -        & 100.0 / 0.13  & 99.93 / 0.26 & - / -        \\
      JCLSR~\cite{zhao2020joint}           & 98.94 / -    & 98.17 / -    & - / -         & - / -        & - / -        \\
      \whline{0.1pt}
      ArcFace~\cite{deng2019arcface} + MB           & 97.92 / 0.009    & 98.73 / 0.012    & 98.58 / 0.014         & 98.83 / 0.008        & 96.12 / 0.022        \\
      ArcFace~\cite{deng2019arcface} + MB + Ours           & \CC{15}99.75 / 0.004    & \CC{15}100.0 / 0.000    & \CC{15}100.0 / 0.000         & \CC{15}100.0 / 0.000 & \CC{15}99.96 / 0.001 \\
   \end{tabular}}
   \vspace{4pt}
   \caption{
      Top-1 accuracy and EER under the `closed-set' protocol.
      Our method significantly improves the top-1 accuracy and EER with a clear margin.
   }\label{tab:closed-set}
   \vspace{-20pt}
\end{table}
\vspace{-15pt}
\mysubsection{Cross-dataset Validation}\label{sec:cross-dataset}
We perform cross-dataset validation to test the generalization of
the proposed method.
We train our method, as well as the baseline (ArcFace),
on one dataset and test the performance on the other dataset.
We test 5 different cross-dataset settings using the MobileFaceNet backbone,
results are summarized in~\cref{tab:cross-dataset}.
The performance is evaluated in terms of both TAR@FAR, EER.

\begin{table}[!htb]
   \centering\vspace{-4pt}
   \setlength\tabcolsep{1.3mm}
   \renewcommand{\arraystretch}{1.2}
   \newcommand{\CC}[1]{\cellcolor{gray!#1}}
   \resizebox{0.65\linewidth}{!}{
   \begin{tabular}{cl|ccc|cc}
   \multirow{2}{*}{Datasets} & \multirow{2}{*}{Method} & \multicolumn{3}{c|}{TAR@FAR=} \\
    & & 1e-3 & 1e-4 & 1e-5 & Top-1 & EER \\
   \whline{1.2pt}
   \multirow{2}{*}{M$\rightarrow$P} & AF   & 0.9759 & 0.9499 & 0.9210 & 99.93 & 0.007 \\
                                    & Ours & \CC{15}0.9935 & \CC{15}0.9766 & \CC{15}0.9622 & \CC{15}100.0 & \CC{15}0.002 \\
   \multirow{2}{*}{T$\rightarrow$P} & AF   & 0.9347 & 0.8981 & 0.8509 & 98.22 & 0.018 \\
                                    & Ours & \CC{15}0.9918 & \CC{15}0.9748 & \CC{15}0.9591 & \CC{15}100.0 & \CC{15}0.003 \\
   \multirow{2}{*}{I$\rightarrow$P} & AF   & 0.9364 & 0.9001 & 0.8020 & 97.67 & 0.019 \\
                                    & Ours & \CC{15}0.9688 & \CC{15}0.9224 &   \CC{15}0.8728    & \CC{15}99.04 & \CC{15}0.009\\
   \multirow{2}{*}{T$\rightarrow$I} & AF   & 0.8533 & 0.7872 & 0.7306 & 97.47 & 0.033 \\
                                    & Ours  & \CC{15}0.9896 & \CC{15}0.9864 & \CC{15} 0.9745   & \CC{15}98.85 & \CC{15}0.007\\
   \multirow{2}{*}{M$\rightarrow$I} & AF   & 0.9927 & 0.9846 & 0.9717  & 99.76 & 0.004 \\
                                    & Ours & \CC{15}1.0000 & \CC{15}1.0000 &   \CC{15}1.0000    & \CC{15}100.0 & \CC{15}0.000 \\
   \end{tabular}
   }
   \vspace{6pt}
   \caption{
      Cross-dataset validation.
      `M', `P', `T' and `I' represent MPD, PolyU, TCD, and IITD datasets, respectively.
      M$\rightarrow$P indicates the model is trained on M and evaluated on P.
   }\label{tab:cross-dataset}
   \vspace{-26pt}
\end{table}

As shown in~\cref{tab:cross-dataset},
our method consistently improves the performance of ArcFace
on all the 5 settings,
suggesting strong cross-dataset generalization ability.

\mysubsection{Palmprint Recognition at Million Scale}\label{sec:million}
To verify the scalability of our proposed method, we test our method
on our internal dataset with million samples.
%
The training set contains 19,286 identities and 2.87 million samples,
while the test set has 1,000 identities and 0.18 million samples.
The images of the dataset are collected parallelly in three
places by 19 difference mobile phones (different brands and modes) and 2 IoT cameras.
Images of each identity was collected in one seesion by 4 devices
(2 IoT and 2 random mobile phones) and 4 different man-made light conditions.
More detailed information and example images of this dataset can be found in ~\cref{appendix:million-scale}.

\begin{figure}[!htb]
   \centering
   \begin{overpic}[width=0.7\linewidth]{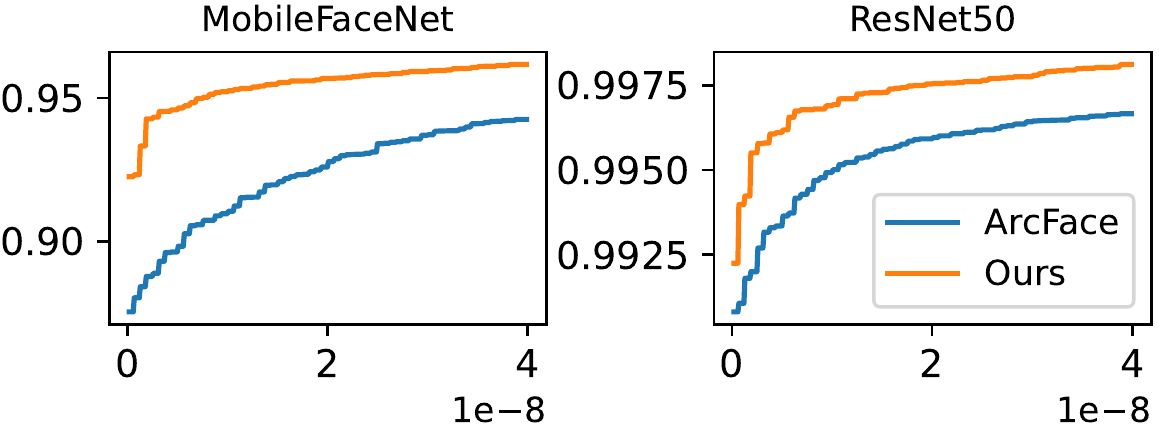}
   \end{overpic}
   \vspace{-10pt}
   \caption{
      FAR \emph{v.s.} TAR curves of ArcFace (AF) and our method
      on the million-scale dataset.
   }\label{fig:roc-1m}\vspace{-6pt}
\end{figure}

We synthesize 20,000 identities and totally 2 million samples to pretrain
the models in this experiment.
The performance is evaluated under open-set protocol
and we report both TAR@FAR and TAR \emph{v.s.} FAR curves in~\cref{tab:million} and~\cref{fig:roc-1m}, respectively.
The results show that our method consistently improves the performance of the baseline
ArcFace method, showing great potential in large-scale palmprint recognition.
\ifdefined \arXiv
\begin{table}[!htb]
\else
\begin{wraptable}{r}{0.65\textwidth}
\fi
   \centering
   \setlength\tabcolsep{1mm}
   \renewcommand{\arraystretch}{1.1}
   \newcommand{\CC}[1]{\cellcolor{gray!#1}}
   \vspace{-20pt}
   \resizebox{0.6\linewidth}{!}{
   \begin{tabular}{lc|cccccc}
   \multirow{2}{*}{Method} & \multirow{2}{*}{Backbone} & \multicolumn{6}{c}{TAR@} \\
   & & 1e-5 & 1e-6 & 1e-7 & 1e-8 & 1e-9 \\
   \whline{1.2pt}
   AF~\cite{deng2019arcface}        & \multirow{2}{*}{MB}    & 0.9911 & 0.9770 & 0.9550 & 0.9251 & 0.8833 \\
   Ours   &                                                  & \CC{15}0.9934 & \CC{15}0.9803 & \CC{15}0.9605 & \CC{15}0.9301 & \CC{15}0.9015\\
   AF~\cite{deng2019arcface}         & \multirow{2}{*}{R50} & 0.9997 & 0.9986 & 0.9964 & 0.9931 & 0.9879 \\
   Ours   &                                                  & \CC{15}0.9999 & \CC{15}0.9996 & \CC{15}0.9975 & \CC{15}0.9943 & \CC{15}0.9911 \\
   \end{tabular}
   }
   \caption{
      Palmprint recognition performance on million scale dataset.
   }
   \label{tab:million}
\ifdefined \arXiv
\end{table}
\else
\end{wraptable}{r}{0.65\textwidth}
\fi

\mysubsection{Palmprint Recognition with Limited Identities}\label{sec:limited-identities}
The model performance under a limited number of training identities is critical to
privacy-sensitive conditions where collecting training set with large-scale identities
is infeasible.
Here we test our method with various training identities.
Specifically, under the open-set protocol (train:test = 1:1), we fix the test set
and train models with 400, 800 and 1,600 identities.
\vspace{-15pt}
\begin{table}[!htb]
   \centering
   \setlength\tabcolsep{1.5mm}
   \renewcommand{\arraystretch}{1.2}
   \newcommand{\CC}[1]{\cellcolor{gray!#1}}
   \resizebox{0.65\linewidth}{!}{
   \begin{tabular}{lc|cccc}
   \multirow{2}{*}{Method} & \multirow{2}{*}{\#ID} & \multicolumn{4}{c}{TAR@FAR=} \\
   & & 1e-3 & 1e-4 & 1e-5 & 1e-6 \\
   \whline{1.2pt}
   ArcFace~\cite{deng2019arcface} & \multirow{2}{*}{1,600} & 0.9292 & 0.8568 & 0.7812 & 0.7049 \\
   ArcFace~\cite{deng2019arcface}+Ours   &  &                \CC{15}0.9640 & \CC{15}0.9438 & \CC{15}0.9102 & \CC{15}0.8437 \\
   ArcFace~\cite{deng2019arcface} & \multirow{2}{*}{800} &   0.8934 & 0.7432 & 0.7104 & 0.6437 \\
   ArcFace~\cite{deng2019arcface}+Ours   &  &                \CC{15}0.9534 & \CC{15}0.9390 & \CC{15}0.9025 & \CC{15}0.8164 \\
   ArcFace~\cite{deng2019arcface}  & \multirow{2}{*}{400} &  0.8102 & 0.7050 & 0.6668 & 0.3320 \\
   ArcFace~\cite{deng2019arcface}+Ours   &  &                \CC{15}0.9189 & \CC{15}0.8497 & \CC{15}0.7542 & \CC{15}0.6899 \\
   \end{tabular}
   }
   \caption{
      Performance under various training identities.
      The models are based on the MobileFaceNet backbone.
   }
   \label{tab:limited-identities}
   \vspace{-10pt}
\end{table}
As demonstrated in~\cref{tab:limited-identities}, our method maintains high performance
while the ArcFace baseline degrades quickly as the drop of training identities.
Even trained with 400 identities, our method still performs on par with the ArcFace counterpart
that is trained with 1,600 identities,
showing its superiority in identity-constrained scenarios.

\mysubsection{Ablation Study}\label{sec:ablation}
In this section, we ablate the components and design choices of our method.
All the experiments in this ablation study are conducted using the
MobileFaceNet~\cite{chen2018mobilefacenets}
and evaluated under the open-set protocol.

\mypar{Creases synthesis}
The main components in our synthesized samples are the
\begin{wraptable}{r}{0.5\textwidth}
   \centering\vspace{-20pt}
   \resizebox{1\linewidth}{!}{
   \renewcommand{\arraystretch}{1.1}
   \newcommand{\CC}[1]{\cellcolor{gray!#1}}
   \begin{tabular}{ccc|cccc}
      \multirow{2}{*}{P} & \multirow{2}{*}{W} & \multirow{2}{*}{B} & \multicolumn{4}{c}{TAR@FAR=} \\
      & & & 1e-3 & 1e-4 & 1e-5 & 1e-6 \\
      \whline{1.2pt}
      \multicolumn{3}{c|}{Baseline}        & 0.9102 & 0.8259 & 0.7458 & 0.7217 \\
      \checkmark     &           &         & 0.9514 & 0.9003 & 0.7613 & 0.7513 \\
      \checkmark &\checkmark &             & 0.9597 & 0.9307 & 0.8949 & 0.8061 \\
      \checkmark &\checkmark & \checkmark  & \textbf{0.9640} & \textbf{0.9438} & \textbf{0.9102} & \textbf{0.8437} \\
   \end{tabular}}
   \vspace{-10pt}
   \caption{
      Ablation of design choices in our method.
   }\label{tab:ablate-components}\vspace{-20pt}
   \vspace{-6pt}
\end{wraptable}
principal lines,
the wrinkles,
and the background images.
~\cref{tab:ablate-components} presents the results of models with and without
these components.
`P', `W' and `B' represent the principal lines, wrinkles and image background
in the synthesized samples, respectively.

Synthesizing principal lines significantly improves the performance over the baseline
at higher FARs, and the improvements at lower FARs are marginal.
With wrinkles, the performance can be further improved especially at
lower FARs.
Finally, using natural images as the background helps achieve higher performance.

\mypar{Compared to imagenet pretrain}
Many down-stream vision tasks, \eg detection, and segmentation,
strongly rely on the imagenet~\cite{deng2009imagenet} pretrained models.
In this experiment, we compare the performance of our synthetically pretrained models
to the imagenet pretrained models.
We pretrain the MobileFaceNets with the imagenet dataset
and our synthesized samples and compare their performance under
the open-set protocol (train:test = 1:1).
For imagenet pretraining, we follow the training configuration of~\cite{he2016deep}.
It is worth noting that there are 1.2 million images in the imagenet training set
and our synthesized dataset consists of only 0.4 million samples
(4,000 identities with 100 samples per identity).
\begin{wraptable}{r}{0.5\textwidth}
   \centering
   \vspace{-10pt}
   \resizebox{1\linewidth}{!}{
   \newcommand{\CC}[1]{\cellcolor{gray!#1}}
   \begin{tabular}{c|ccccc}
      \multirow{2}{*}{Pretrain} & \multicolumn{4}{c}{TAR@FAR=} \\
      & 1e-3 & 1e-4 & 1e-5 & 1e-6 \\
      \whline{1.2pt}
      Imagenet & 0.9608 & 0.9135 & 0.8294 & 0.7256 \\
      Ours     & \CC{15}0.9640 & \CC{15}0.9438 & \CC{15}0.9102 & \CC{15}0.8437 \\
   \end{tabular}}
   \caption{
      Comparison of imagenet and our synthetically pretrained models.
   }\label{tab:imagenet-pretrain}\vspace{-10pt}
\end{wraptable}
As demonstrated in
~\cref{tab:imagenet-pretrain},
even pretrained with one-third of samples, our proposed method still outperforms
the imagenet pretrained model with a clear margin, especially under lower FARs.
The experimental result tells that our synthesized dataset
is specifically more suitable for palmprint recognition than general
vision datasets, \eg imagenet.

\mypar{Number of synthesized samples and identities}
By default, we synthesize
\begin{figure}[!htb]
   \centering\vspace{-10pt}
   \begin{overpic}[width=0.8\linewidth]{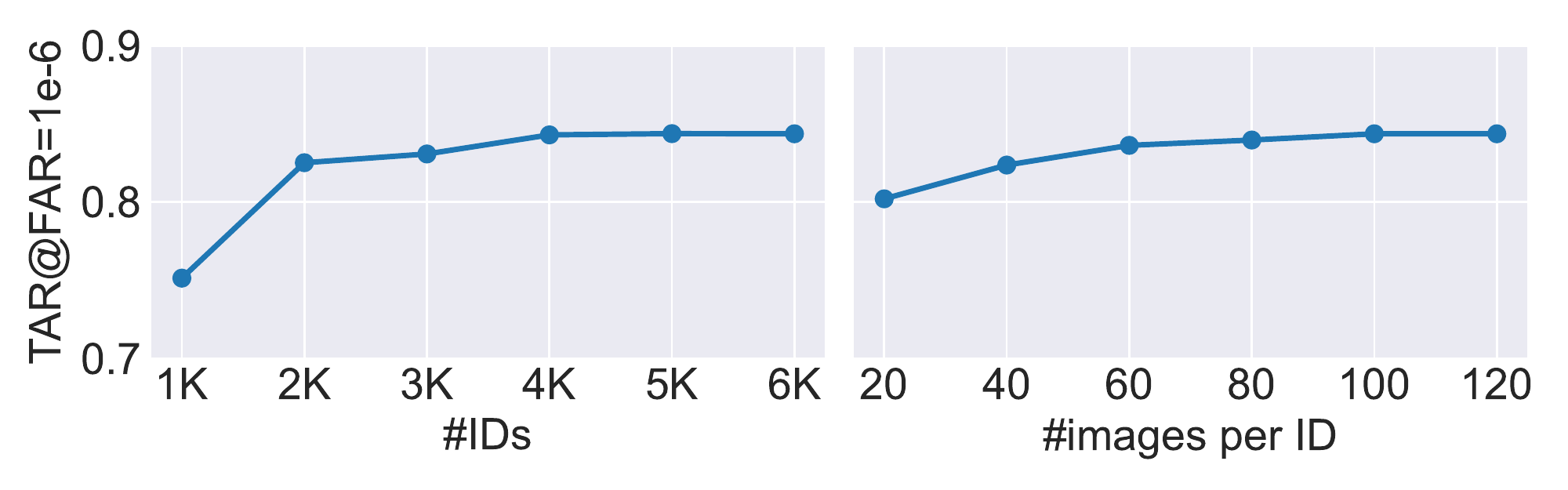}
   \end{overpic}
   \vspace{-10pt}
   \caption{
      TAR@FAR=1e-6 of models pretrained with different synthetic
      samples and identities.
   }\label{fig:ablate-num}
   \vspace{-10pt}
\end{figure}
4,000 identities and each of them has 100 images.
In this ablation, we fix one number as the default
and vary the other, and evaluate the
finetuned performance in terms of FAR@1e-6.
The results in~\cref{fig:ablate-num} reveal that
increasing both the number of samples and identities improves the performance.
The number of identities has a greater impact on the fine-tuned performance
and the number of samples has less impact.


\mysection{Conclusion}
We proposed a simple yet effective geometric model
to synthesize palmar creases by manipulating
parameterized B\`ezier curves.
The synthetic samples are used to pretrain deep palmprint recognition
models and improve model performance.
Different from other data synthesizing methods, our method synthesizes samples
of novel categories to augment both the identities and samples of the training set.
Competitive
results on several public benchmarks demonstrate the
superiority and great potentials of our approach.
Besides, experiments on a million-scale dataset
verify the scalability of our method.
We also believe our method could benefit some other tasks,
~\eg fingerprint recognition.

\ifdefined \arXiv
   \section*{Acknowledgment}\vspace{-10pt}
   We would like to acknowledge Haitao Wang and Huikai Shao for their
   assistance in processing experimental results.
\else
   \clearpage
   \pagebreak
\fi

\bibliographystyle{splncs04}
\bibliography{palm}

\appendix

\section{Details synthesis of principal lines} \label{appendix:synthesize-details}
We elaborate the steps of the synthesizing details using a left hand of 3 principal lines.
\begin{figure}[!htb]
   \centering
   \begin{overpic}[width=0.8\linewidth]{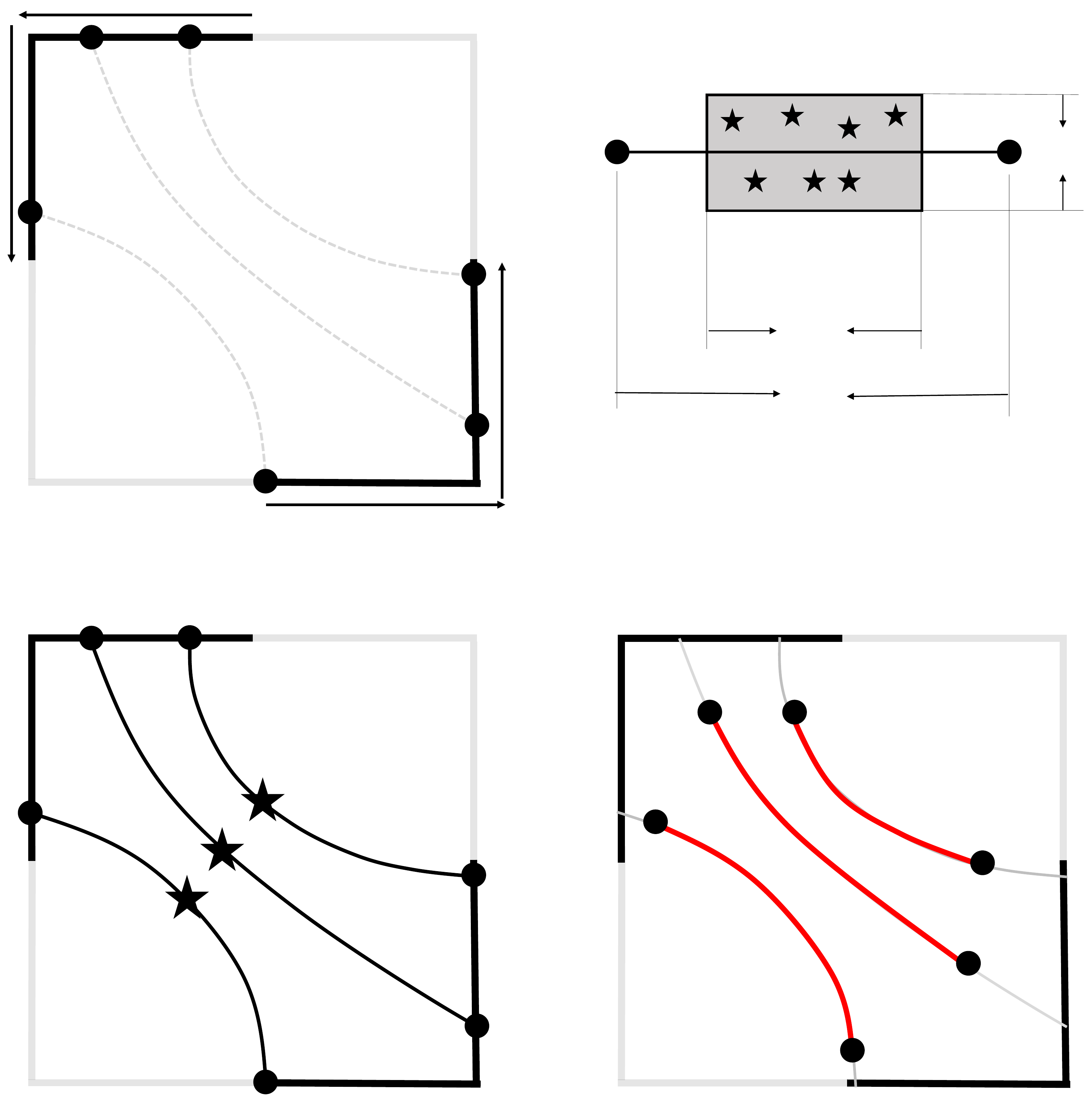}
      \put(22.5,50){(a)}
      \put(76,50){(b)}
      \put(22.5,-4){(c)}
      \put(76,-4){(d)}
      \put(22.5,100){0}
      \put(-1.5,76){1}
      \put(22,53){0}
      \put(45,77){1}
      \put(17,94){$s_1$}
      \put(8,94){$s_2$}
      \put(4,80){$s_3$}
      \put(39,75){$e_1$}
      \put(39,61){$e_2$}
      \put(23,58){$e_3$}
      \put(73,63){$L$}
      \put(73,69){$\frac{2}{3}L$}
      \put(95.4,85.3){$\frac{1}{3}L$}
      \put(18,39){$s_1$}
      \put(40,23){$e_1$}
      \put(26,28){$c_1$}
   \end{overpic}\vspace{10pt}
   \caption{
      Detailed steps of principal lines of a left hand.
   }\label{fig:detailed-steps}
   \vspace{-5pt}
\end{figure}
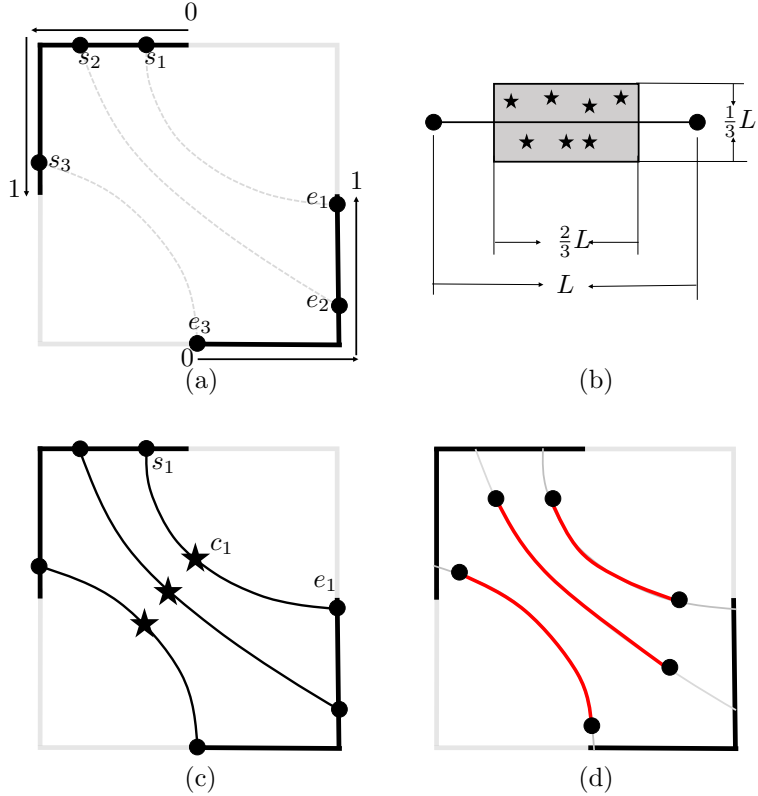
\begin{itemize}\setlength\itemsep{0.5em}
   \item ~\cref{fig:detailed-steps} (a): Randomly select starting points ($s_1, s_2, s_3$) and ending points ($e_1, e_2, e_3$) along the top-left
         and bottom-right edge of the coordinate.
         We set a simple rule to ensure that the lines do not intersect: $S_3 > s_2 > s_1$ and $S_3 > s_2 > s_1$.

   \item ~\cref{fig:detailed-steps} (b): 
         The control points are randomly sampled from a rectangle area.
         Let $L$ be the length of the line between $s$ and $e$, the width and height of the rectangle
         are $\frac{2}{3}L$ and $\frac{1}{3}L$.
   
   \item ~\cref{fig:detailed-steps} (c): 
         B\'ezier curves are determined given the starting, end and control points ($s, e, c$).
         Let $f(t): \mathbb{R}^{[0, 1]}\rightarrow \mathbb{R}^2$ be the parametric function,
         where $s = f(0)$ and $e = f(1)$.

   \item ~\cref{fig:detailed-steps} (d):
         Randomly sample $t_0 \in (0, 0.3)$, $t_1 \in (0.7, 1)$ so that the finally curves ({\color{red}{red}}) are $f(t), t\in[t_0, t_1]$.
\end{itemize}

\section{Algorithmic pipeline of the synthesizing process}\label{appendix:algorithm}
~\cref{alg:random-generate} illustrates the pipeline of the synthesizing process.
$N$ and  $S$ represent the number of identities and number of samples per identity.
The between identity randomness and within identity randomness are represented by
different collors.
\begin{algorithm}[!htb]
   \setstretch{0.8}
   \caption{Algorithmic pipeline for creases synthesis}\label{alg:random-generate}
   \begin{algorithmic}[1]  
   \For{$i\in \{1, 2, .., N\}$}
   \State {\color{blue}{\texttt{m = randint(3, 5)}}}
   \State {\color{blue}{\texttt{n = randint(5, 15)}}}
   \State {\color{blue}{\texttt{$P$ = random(0, 1, size=(m, 3, 2))}}}
   \State {\color{blue}{\texttt{$Q$ = random(0, 1, size=(n, 3, 2))}}}
      \For{$j \in \{1, 2, .., S\}$}
         \State {\color{ao(english)}{$P^i_j$ += random($P$, std=0.04)}}
         \State {\color{ao(english)}{$Q^i_j$ += random($Q$, std=0.01)}}
         \State {\color{ao(english)}{bg = random\_select(imagenet)}}
         \State $S^i_j = \text{synthesize}(P^i_j, Q^i_j, bg)$
      \EndFor
   \EndFor
   \end{algorithmic}
\end{algorithm}

\section{Example of synthesized images} \label{appendix:example}
\begin{figure}[!htb]
   \centering
   \begin{overpic}[width=0.6\linewidth]{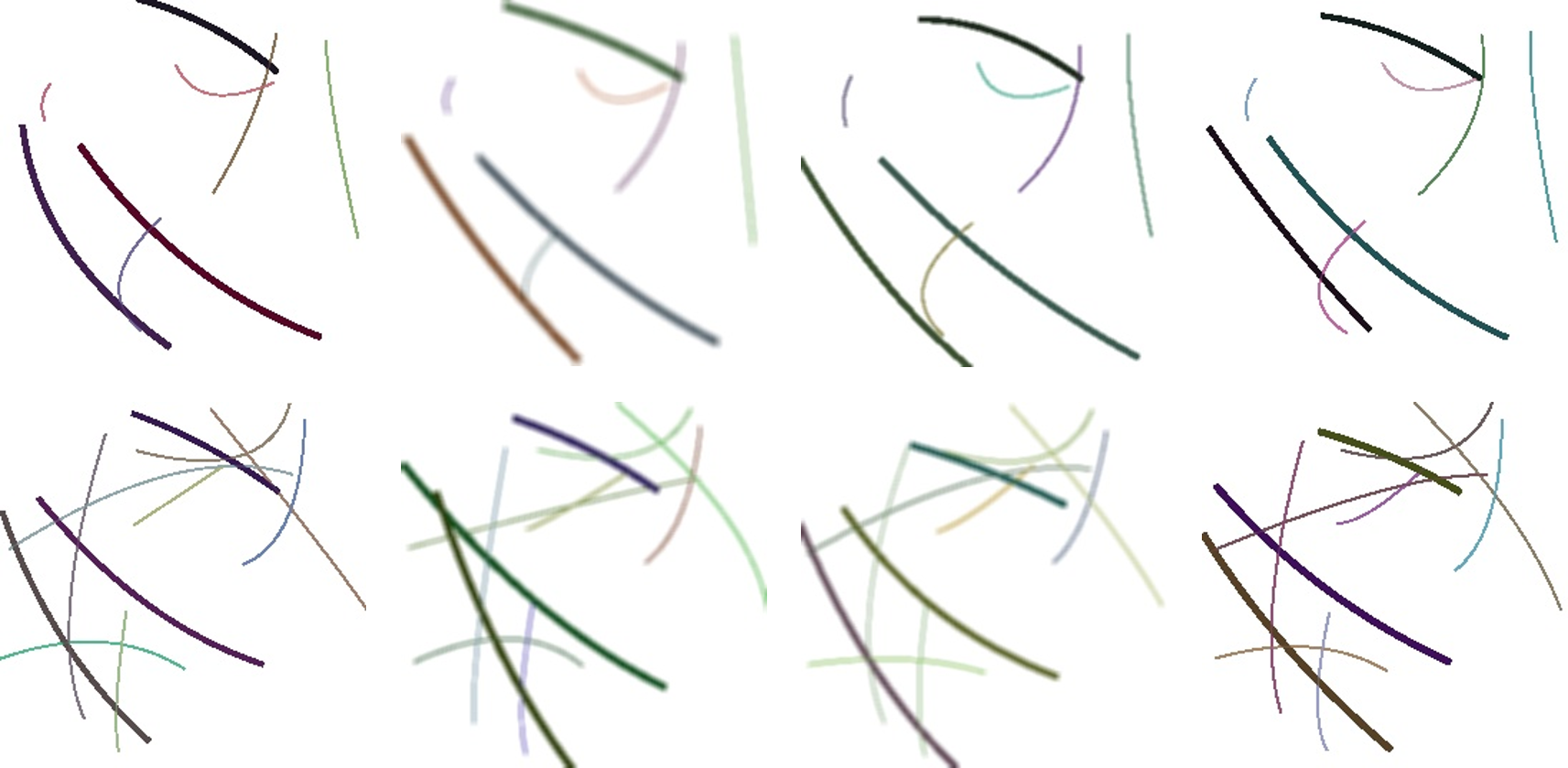}
   \end{overpic}\vspace{-5pt}
   \caption{
      Example of synthesized images without imagenet images as background.
      Each row contains sample of the same identity.
   }\label{fig:example-synthesis}
\end{figure}

\begin{figure}[!htb]
   \centering
   \begin{overpic}[width=0.6\linewidth]{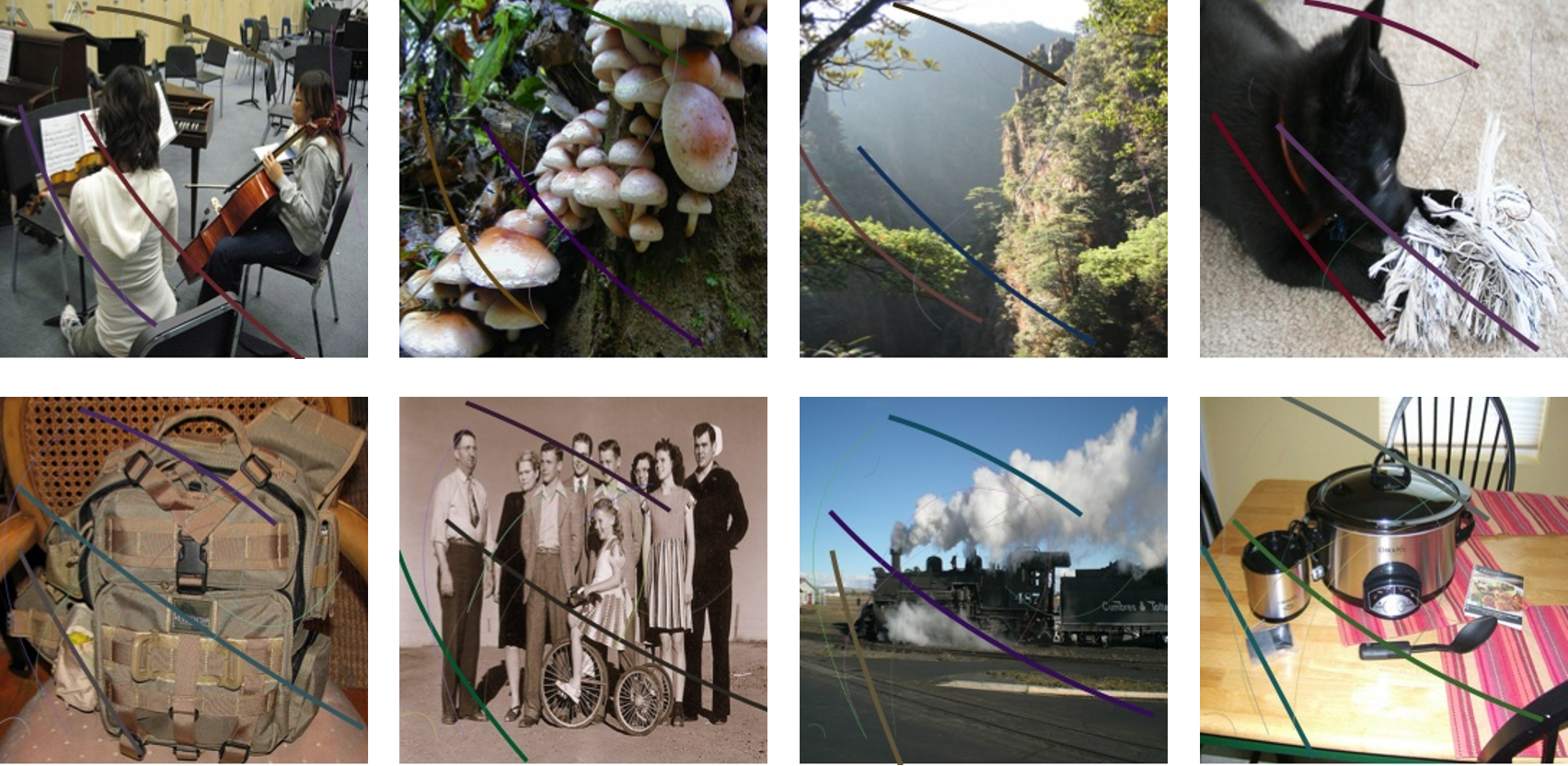}
   \end{overpic}\vspace{-5pt}
   \caption{
      Example of synthesized images with imagenet images as background.
      Each row contains sample of the same identity.
   }\label{fig:example-synthesis-bg}
\end{figure}

~\cref{fig:example-synthesis} and ~\cref{fig:example-synthesis-bg} present example synthesized images
without and with imagenet images as the background.
\vspace{-20pt}

\section{Details about the million-scale dataset} \label{appendix:million-scale}
The images of the the dataset are collected parallelly in three
places by 19 difference mobile phones (different brands and modes) and 2 IoT cameras.
Images of each identity was collected in one seesion by 4 devices
(2 IoT and 2 random mobile phones) and 4 different man-made light conditions.
\begin{figure}[!htb]
    \centering
    \vspace{-1.5em}
    \includegraphics[width=0.9\linewidth]{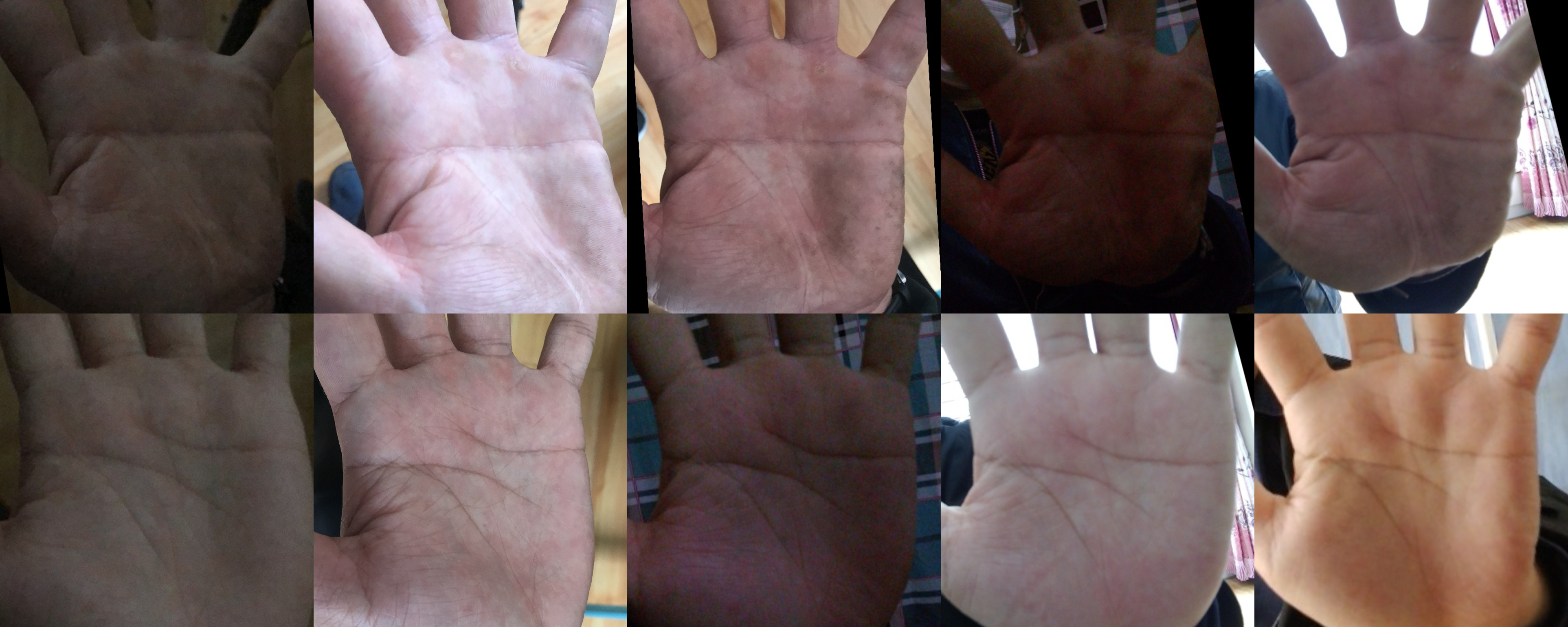}
    \vspace{-10pt}
    \caption{
       Example images of two identities (each row corresponds to an identity) our million-scale dataset.
    }
\end{figure}
We provide selected palms of two identities in the figure below (zoom in for details).
\end{document}